\documentclass[letterpaper, 10 pt, conference]{ieeeconf}  

\IEEEoverridecommandlockouts                              

\usepackage{graphicx} 
\usepackage{amsmath} 
\usepackage{amssymb}  

\graphicspath{ {./img/} }

\title{\LARGE \bf
Learning a Directional Soft Lane Affordance Model for Road Scenes Using Self-Supervision
}

\author{Robin Karlsson$^{1*}$ and Erik Sjoberg$^{2}$
\thanks{$^{1}$Robin Karlsson (corresponding author) is a researcher at Tier IV, Tokyo, Japan.
        {\tt\small robin.karlsson@tier4.jp}}%
\thanks{$~{*}$ This work was done while employed at Ascent Robotics}
\thanks{$^{2}$Erik Sjoberg is a senior research engineer at Ascent Robotics, Tokyo, Japan.
        {\tt\small erik@ascent.ai}}%
}

\begin{document}

\maketitle
\thispagestyle{empty}
\pagestyle{empty}

\begin{abstract}

Humans navigate complex environments in an organized yet flexible manner, adapting to the context and implicit social rules. Understanding these naturally learned patterns of behavior is essential for applications such as autonomous vehicles. However, algorithmically defining these implicit rules of human behavior remains difficult. This work proposes a novel self-supervised method for training a probabilistic network model to estimate the regions humans are most likely to drive in as well as a multimodal representation of the inferred direction of travel at each point. The model is trained on individual human trajectories conditioned on a representation of the driving environment. The model is shown to successfully generalize to new road scenes, demonstrating potential for real-world application as a prior for socially acceptable driving behavior in challenging or ambiguous scenarios which are poorly handled by explicit traffic rules.

\end{abstract}

\section{INTRODUCTION}

Humans tend to navigate complex environments restricted by rules and obstacles in an organized and predictable manner, following a pattern according to the environment context such as city streets and factory floors \cite{krause13}. One example of a movement pattern is the concept of directional lanes, which is a fundamental part of how traffic is organized, enabling the road to be shared by many independent traffic participants. Being able to infer and follow mutual lanes is a critical feature of an autonomous vehicle intended to share the road with human drivers \cite{urmson08}. Humans are able to implicitly infer road lanes purely by observing drivable road, road markings, signs, and obstacles, both in structured and semi-unstructured roads such as intersections, parking lots, and roads without distinct markings.

While many current high-profile autonomous vehicle projects rely on high-definition (HD) maps of the road for knowing how to follow conventional traffic rules and road lanes \cite{sheif16}, humans tend to not respect road lanes as hard, discrete concepts, but as guidelines for complex multi-agent interaction. The correct traffic behavior also depends on local conventions and rules, and thus needs to be adjusted for every region.

Deducing a mathematical model for correct driving behaviour given an arbitrary road scene is non-trivial. However, driving data is abundant and can be utilized by machine learning methods to learn to understand the intent of a roadway given unclear, ambiguous or even contradictory markings and signs, as well as obstacles and debris. This can allow the model to extract behaviour perceived as natural to humans. 

We define the concepts Directional Affordance (DA) and Soft Lane Affordance (SLA) in order to represent directional road lanes. SLA is defined as the belief that a grid map element belongs to a road lane. DA is a probabilistic multimodal distribution expressing directionality at each element.

\begin{figure}[t]
  \centering
  \includegraphics[width=0.45\textwidth]{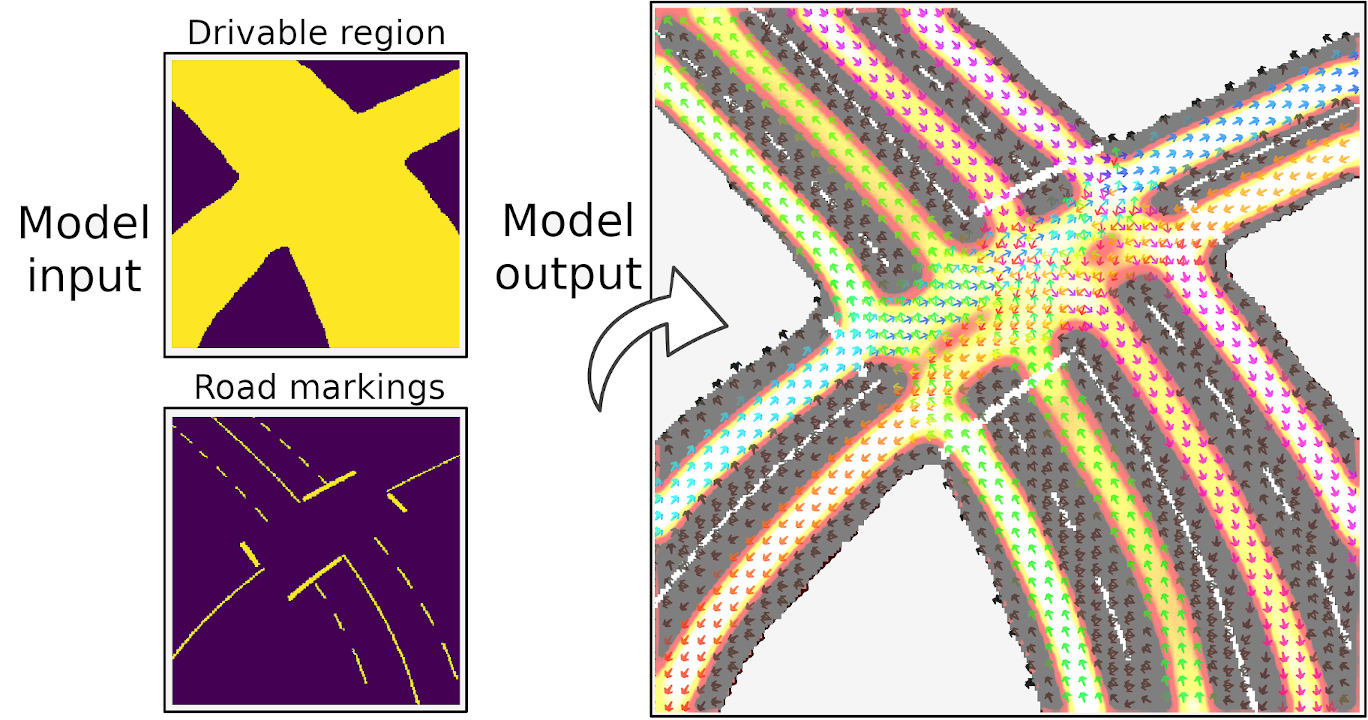}
  \caption{The model predicts directional road lanes based on an input representing the drivable region and road markings of a road scene.}
  \label{fig:model_input_output}
\end{figure}

\begin{figure}
  \centering
  \includegraphics[width=0.35\textwidth]{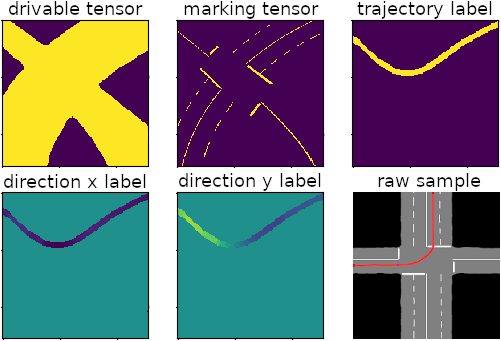}
  \caption{Visualization of an augmented training sample as supplied to the model. The drivable region and road marking tensors correspond to the two input layers. The three label tensors consists of the example trajectory with directional $(\hat{n}_x, \hat{n}_y)$ unit vector components. The bottom right image shows the unaugmented raw sample with the example trajectory.}
  \label{fig:training_sample}
\end{figure}

The main contribution of this paper is a self-supervised method to learn a probabilistic network model for inferring all feasible directional paths or road lanes given the context of the roadway in terms of SLA and DA, as illustrated in Fig.~\ref{fig:model_input_output}. The model is trained on single trajectory examples as shown in Fig.~\ref{fig:training_sample}. The training process can be automated and is suitable for online learning. The proposed model displays feature independence, and is shown to generalize to new intersections and asymmetric road layouts. The SLA and DA output can be utilized in the form of a dense cost map or post-processed into discrete road lane waypoints \cite{salzmann19}.

The propsed model can be used as an HD-map subsitute, which illustrates its significance in autonomous driving technology. When a map is unavailable or when localization errors occur, this method can allow urban road navigation with on-board sensor information only. Additional usages are consistency evaluation of the onboard map given the actual observed environment, as well as a tool for map generation.

The rest of the paper is organized as follows. Sec.~\ref{fig:related_work} summarizes the state of the art. Sec.~\ref{sec:dsla} explains the proposed model as well as its input and output. The training process including loss term formulations is explained in Sec.~\ref{sec:training_process}. The experiment setup, data augmentation, and performance evaluation metrics, are explained in Sec.~\ref{sec:experiments}. The performance, learning characteristics, and failure cases of the proposed model are evaluated in Sec~\ref{sec:results}. Sec.~\ref{sec:conclusion} concludes the paper by summarizing the results and suggests future work.

\section{RELATED WORK}
\label{fig:related_work}

\subsection{Path prediction}
Recent studies have presented models trained to output multimodal path predictions for specific actors. In particular Salzmann et al.~\cite{salzmann19} trained a probabilistic convolutional neural network (CNN) model to output a dense grid of elements representing road lanes conditioned on a driving direction. The model is trained through a pixel-wise classification error with a weighted mask loss on automatically recorded driving trajectories and top-down road scene representations consisting of features such as road geometries and road markings similar to our work. The resulting path map is integrated with an MPC controller to demonstrate trajectory planning  as well as used to validate an on-board HD map during operation. Other related studies include Baumann et al.~\cite{baumann18} who trained a CNN model to output path predictions for single actors using human driving examples. The loss function is masked in order to balance the amount of predicted path and not-path grid elements. Pérez-Higueras et al.~\cite{perezhigueras18} presented a CNN model to infer a multimodal path between two points. Training labels consist of a human-generated example trajectory connecting the two points. The model output is used as a prior by an RRT* path planner \cite{karaman11} to generate trajectories for navigation. Barnes et al.~\cite{barnes17} trained a multimodal path prediction CNN model using self-supervised learning, where training samples are generated automatically from driving data. Kitani et al.~\cite{kitani12} trained a Hidden Parameter Markov Decision Process (HiP-MDP) model~\cite{doshivelez12} based on an inverse reinforcement learning approach~\cite{abbeel04} using human observation data. Ratliff~\cite{ratliff09} learned a mapping from input map features to cost using supervised learning. An optimal policy MDP following the learned cost will result in trajectories which mimic the expert's behavior.

Our work extends on these studies by presenting a model which not only predicts actor-specific paths, but generalizes to output actor-independent paths across the entire road scene, as well as inferring a multimodal probabilistic directionality necessary for distinguishing intersecting paths. Additionally, we propose a quantitative performance evaluation metric to fairly evaluate actor-specific path and road lane prediction methods, which is lacking in \cite{salzmann19}.

\subsection{Semi-automatic HD map generation}

Recent studies include Iesaki et al.~\cite{iesaki19} who presented a method to connect intersection roadways with polynomial curves fitted according to a cost function outputted by a learned model. However, the model relies on a priori knowledge on how lanes are connected and thus the method is not generalizable to new road layouts. Guo et al.~\cite{guo16} used human driving data and a geometric intersection model to fit clothoids. The model does not learn how intersection roadways are connected though contextual features, and thus cannot generalize to new roadways lacking human driver data. Additionally, the model relies heavily on road markings and is thus heavily dependent on a particular feature. Zhao et al.~\cite{zhao19} used human driving data with a SLAM-based approach to generate a vectorized road lane map. These semi-automatic HD map generation methods do not generalize by learning a model of the road network through contextual features, and thus cannot be applied to new environments unlike our proposed method.

\subsection{Scene understanding}

Related works include Wang et al.~\cite{wang19} who generated a semantic representation of the road from front-view images, without explicitly inferring directionality or lanes. Kunze et al.~\cite{kunze18} semantically connected road lanes as a scene graph, without spatially anchoring the nodes in the road scene. Geiger et al.~\cite{geiger11} presented a model which infers the topology, geometry, and directionality of the road scene. However, the fidelity of the output is low, and thus not suitable for practical application unlike our proposed model.

\section{DIRECTIONAL SOFT LANE AFFORDANCE MODEL}
\label{sec:dsla}

The Directional Soft Lane Affordance (DSLA) model is a probabilistic von Mises mixture density network \cite{bishop04} and a dense affordance map model.

The directional mixture model $p(\theta)$ corresponds to a multimodal probabilitity distribution (\ref{eq:mixture_model}) of road directionality $\theta$ at each grid map element. The model consists of $m \in 1, ..., M$ von Mises distributions $p(\theta)_m$ and mixture weights $w_m$. Each distribution (\ref{eq:vonmises}) is similar to a Gausian distribution with mean angle $\mu_m \in (0, 2 \pi)$ and concentration parameter $b_m$ but being periodic in the domain $\theta \in (0, 2 \pi)$.

\begin{equation} \label{eq:mixture_model}
 p( \theta ) = \sum_m w_m \: p( \theta )_m
\end{equation}

\begin{equation} \label{eq:vonmises}
 p( \theta )_m = \frac{1}{2 \pi I_0(b_m)} exp(b_m  \: cos(\theta - \mu_m))
\end{equation}

The distribution is normalized using the zeroth order Bessel function of the first kind, which is numerically approximated as

\begin{equation} \label{eq:bessel}
 I_0 (b_m) = \frac{1}{2 \pi} \int_0^{2 \pi} exp( b_m cos \theta ) d \theta
\end{equation}

The outputs are generated by a CNN model which is trained with a self-supervised learning method \cite{nava19}. The model is trained on pairs of segmented top-down context images representing aspects of the environment in terms of a grid map encoded with drivable region and road markings, and one example trajectory starting and ending outside the image. As the automatically generated labels only contain one of many feasible trajectories in the road scene, the training method is also relatable to learning from partially labeled data using weak supervision \cite{zhou17}.

Discretizing the input contextual map into grid points, the model outputs a binomial likelihood estimate $Y \in [0, 1]$ of the grid point being part of a soft road lane, and a multimodal directional distribution $p( \theta ), \theta \in [0, 2 \pi]$ for each grid point. In this work, the model is trained to output three distributions to represent directionality for all experiments. This is the minimum required number of distributions required to express the directionality as found by superimposing all example trajectories for each road layout. However, more distributions can be obtained by modifying the model or from sampling mixture components \cite{prokudin18}. Each directional mode $\theta_m$ in the mixture distribution $p(\theta)$ corresponds to a directional mode of the traffic at every grid point.

\subsection{Input and output representation}

The input road context image consists of a 2 layer 256x256 grid array representing an intersection-sized segmented top-down view of the road. The input representation can be generated in real-time using onboard sensors \cite{seo19, salzmann19}. In this work, the road scene is segmented into a drivable layer and a road marking layer, but other contextual information like semantic traffic rules and free-space can be spatially encoded into layers. The segmentation can be encoded probabilistically by assigning true observations as 1.0, unknown elements as 0.5, negative observation as 0.0, with intermediate values encoding observational belief.

The model output consists of a 10 layer 128x128 grid array. The first layer corresponds to the soft lane affordance output $Y$. The following layers correspond to sets of directional affordance outputs, each representing the parameters of a von Mises mixture density distribution (\ref{eq:vonmises}); a normalized directional mode $\tilde{\mu}_m \in [0, 1]$, normalized directional variance $\tilde{\sigma} \in [0, 1]$, and unnormalized mixture weight $\tilde{w}_m \in [0, 1]$. Fig.~\ref{fig:train_visualization} shows an example of how the output can be visualized.

\subsection{Model architecture}

\begin{figure}
  \centering
  \includegraphics[width=0.48\textwidth]{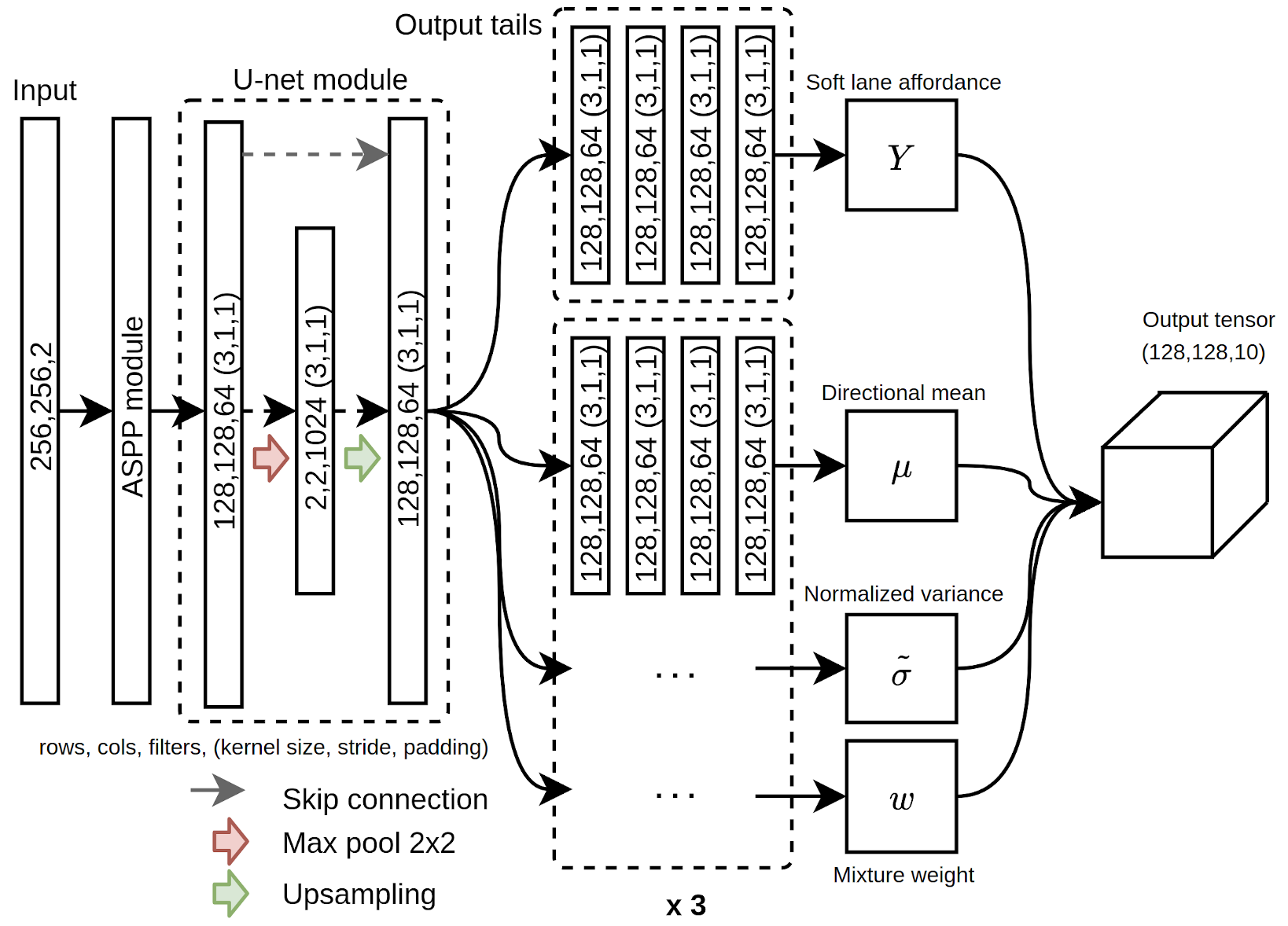}
  \caption{The network module consists of an ASP module \cite{chen16} which extracts features from the input. A modified U-net module \cite{ronneberger15} further processes the features into a shared latent representation. Four task-specific convolutional layers \cite{caruana93} are used to generate each output.}
  \label{fig:model}
\end{figure}

The model architecture is shown in Fig.~\ref{fig:model}. The input features are first processed by an atrous spatial pyramid pooling (ASPP) module \cite{chen16} using eight parallel convolutional layers, which extracts input features at multiple scales and reduces the layer size. The extracted features are fed into a modified U-Net network \cite{ronneberger15} which is a 26 layer encoder-decoder architecture with skip connections to improve contextual reasoning. A variation of this architecture is also used by Salzmann et al.~\cite{salzmann19}. The 2x2 sized bottleneck is important for allowing convolution over the entire latent layers with a 3x3 kernel. The shared latent representation is fed to task-specific layers for each output type \cite{caruana93}. The sigmoid function is used for all output layers. The resulting layers are finally concatenated into a three-dimensional tensor. Nearest neighbour upsampling is found to provide smoother, more continuous output compared with max pooling, as well as eliminating checkerboard patterning \cite{odena16}. The directional output is transformed to von Mises distribution parameters (\ref{eq:vonmises}) by normalizing the mixture weights $w_m = \tilde{w}_m / \sum_{j=1}^M \tilde{w}_j$, rescaling the directional means $\mu_m = 2 \pi \: \tilde{\mu}_m$, and transforming the normalized variances to concentration parameters $b_m = b_{max} (1 - \tilde{\sigma}_m + \epsilon)$. Larger $b_m$ values generate peakier distributions. In this work, $b_{max}$ is limited to 88.0 for numerical reasons. The skip connections of U-Net are found to produce superior results and more efficient learning compared to SegNet \cite{badrinarayanan17} which lacks such connections.

\section{TRAINING PROCESS}
\label{sec:training_process}

The model is trained on samples containing a single trajectory label in an inverse reinforcement approach \cite{bagnell15}, to learn a mapping that results in example trajectories having a high likelihood of being represented given the segmented road context input \cite{ratliff09}.

Every training sample is randomly augmented and thus unique, resulting in an online learning process where every sample is drawn from a data generating distribution $p_{data}(x, y)$. Given that $p_{data}(x, y)$ has sufficient coverage of the true data distribution $p_{true}(x, y)$ \cite{goodfellow16}, that contains a wide variety of general road layouts from which to generate augmented samples, it is guaranteed that the learning gradient will optimize the model to generalize beyond particular road scene geometries, and generate valid output for geometrically different but semantically similar road scenes as demonstrated in Sec.~\ref{sec:results}.

The training process itself consists of a multi-objective loss function consisting of two parts; a soft lane affordance loss term $L_{SLA}$ (\ref{eq:sla_loss}), and a directional affordance loss term $L_{DA}$ (\ref{eq:da_loss}). The total loss is computed as $L_{tot} = \tilde{L}_{SLA} + \tilde{L}_{DA}$. The magnitude of the two losses are normalized according to

\begin{equation} \label{eq:loss_normalization}
 \tilde{L}_{SLA} = L_{SLA} \times L_{DA}^{detached}, \: \: \: \tilde{L}_{DA} = L_{DA} \times L_{SLA}^{detached}
\end{equation}

The reason for normalization is that it is experimentally observed that the two objectives having similar loss function magnitudes allows the model to learn better about each task in parallel, while not requiring loss term weight hyperparameter studies or other more sophisticated multi-task learning methods \cite{kendall18}. It is believed that loss term normalization avoids smaller gradients from one loss term being dominated by the gradients from other loss terms, and hence avoids training stagnation once the value of the dominating loss term saturates, preventing learning from the other loss terms. Care must be taken to ensure that the normalization does not affect the gradient flow, and only scales the magnitude of the loss in order to decouple the two gradients. In the PyTorch framework, the secondary loss term in each multiplication needs to be explicitly detached from the gradient flow.

\subsection{Soft lane affordance loss formulation}

The soft lane affordance loss $L_{SLA}$ is computed for every output grid point $(i, j)$ according to

\begin{equation} \label{eq:sla_loss}
 L_{SLA} = \sum_{i,j} | y_{i,j} - \hat{y}_{i,j} |^2 + \alpha_{SLA} \: \beta \: \sum_{i,j \in mask} | y_{i,j} - \hat{y}_{i,j} |^2
\end{equation}

where $y_{i,j} \in Y$ represents the soft lane affordance model output, $\hat{y}_{i,j} \in \hat{Y}$ represents the label trajectory, $\alpha_{SLA}$ is a scaling term, $\beta = n^2 / \tilde{n}$ is a normalization term where $n^2$ is the total number of grid points in the entire context tensor, and $\tilde{n}$ is the number of grid points masked by the label trajectory. The first term is the overall loss term computed over every element in the output layer. The term will inevitably contain false negatives corresponding to true lane elements which are not encompassed by the single trajectory label. The purpose of the second term is to increase the importance of true soft lane elements by applying a masked loss term computed only over the elements encompassed by the trajectory label. The relative loss weighting of the masked elements is set by $\alpha_{SLA}$. A low value will encourage sparse output as the false negatives will have a larger contribution to the overall loss. It is critical to make the mask loss invariant to the size of the actual masking region, in order to ensure that the model learns to output soft lane affordance with equal importance for long and short soft lanes across the entire output region.

The choice of a mean square error (MSE) loss over a cross entropy (CE) loss is because MSE is more robust to the label noise associated with the false negative labels in $\hat{Y}$ \cite{manwani12}.

\subsection{Directional affordance loss formulation}
\label{sec:da_loss}

The model outputs a multimodal von Mises distribution $p(\theta)$ (\ref{eq:mixture_model}) for every grid point $(i,j)$. The distribution consists of a predetermined or theoretically infinitely sampled number of mixture components $(\mu, \tilde{\sigma}, w)_m$ \cite{prokudin18}.

To compute the directional loss value, the ideal target von Mises distribution $\hat{p} (\theta)_{i,j}$ is computed at every grid point encompassed by the path label. The encoded directional unit vectors $(\hat{n}_x, \hat{n}_y)$ of the directional label elements are used to compute a target mean direction $\hat{\mu}_{i,j}$ for each grid point. The ideal directional output corresponds to $\hat{\mu}_{i,j}$ with minimum normalized variance $\tilde{\sigma} = 0$, which amounts to the maximal concentration parameter $b_{max}$. The ideal target distribution thus becomes

\begin{equation} \label{eq:vonmises_target}
 \hat{p}(\theta)_{i,j} = \frac{1}{2 \pi I_0(b_{max})} exp ( b_{max} \: cos (\theta - \hat{\mu}_{i,j}) )
\end{equation}

KL divergence (\ref{eq:kl_div}) is used as the directional loss function. The loss decreases towards zero as the output distribution $p(\theta)_{i,j}$ approaches the ideal target distribution $\hat{p}(\theta)_{i,j}$.

\begin{equation} \label{eq:kl_div}
 D_{KL} (\hat{p}(\theta)_{i,j} \| p(\theta)_{i,j}) = \int_0^{2 \pi} \hat{p}(\theta)_{i,j} ln \frac{\hat{p}(\theta)_{i,j}}{p(\theta)_{i,j}} d \theta 
\end{equation}

The directional loss $L_{DA}$ is computed by taking the mean of (\ref{eq:kl_div}) over all $\tilde{n}$ grid points encompassed by the path label.

\begin{equation} \label{eq:da_loss}
 L_{DA} = \frac{1}{\tilde{n}} \sum_{i,j} D_{KL}( \hat{p}(\theta)_{i,j} \| p(\theta)_{i,j} )
\end{equation}

The reason for choosing a KL divergence loss over a pointwise expectation maximization loss is because simple calculation shows that the former loss favors a multimodal distribution over a wide monomodal distribution when optimized over two separate monomodal distributions. Additionally, the KL divergence allows for potentially training a model using multimodal directional labels.

\section{EXPERIMENTS}
\label{sec:experiments}

Seven models with different hyperparameters are trained and evaluated in terms of learning on samples from the training distribution $p_{data}(x,y)$ and ability to generalize to samples from the test distribution $p_{
test}(x, y)$. The experiments are specified in Table~\ref{tab:summary}. Experimental hyperparameters were chosen around a set of working parameters in order to assess the general influence of each hyperparameter on performance. The model parameters are trained using stochastic gradient descent using the Adam optimizer with a stepped learning rate reduction of $10\%$ every 100 epochs. Single sample training is motivated by the beneficial regularizing effect \cite{wilson04}, potentially better generalization ability \cite{goodfellow16}, and compatibility with computationally heavy online data augmentation.

\subsection{Data}
\label{sec:data}

\begin{figure}
  \centering
  \includegraphics[width=0.45\textwidth]{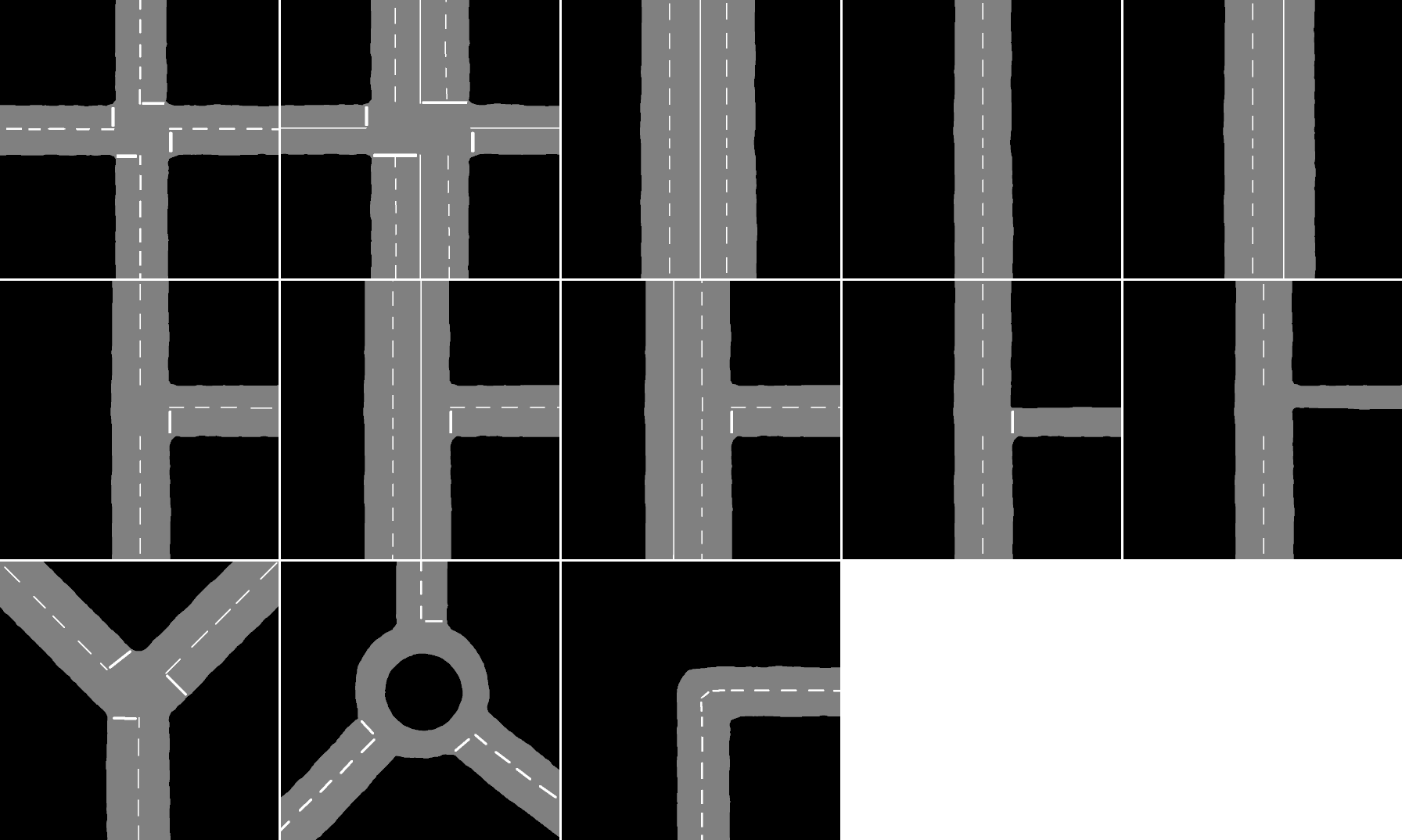}
  \caption{Road layouts representing the training distribution $p_{data}(x, y)$. Road scenes sampled from $p_{data}(x, y)$ are intended to consist of essential structural constituents found in common road scenes.}
  \label{fig:train_dist}
\end{figure}

\begin{figure}
  \centering
  \includegraphics[width=0.45\textwidth]{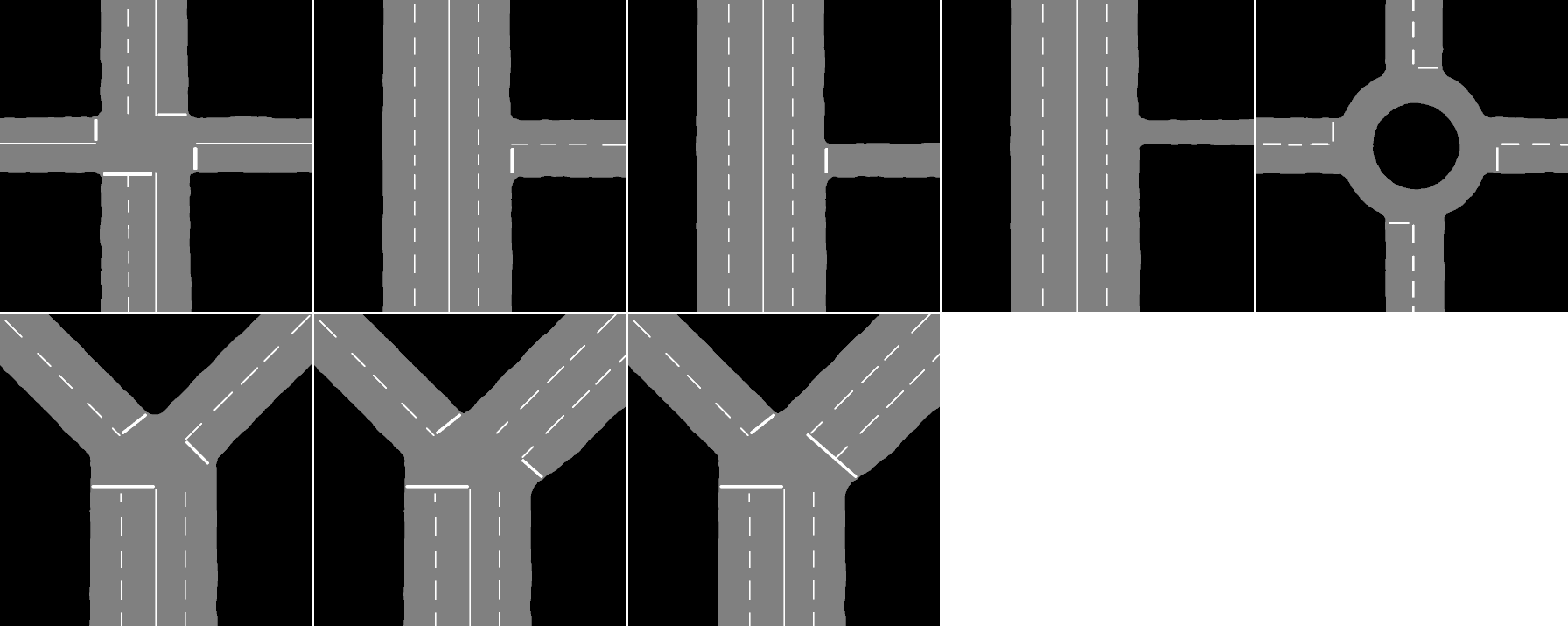}
  \caption{Road layouts representing the testing distribution $p_{test}(x, y)$. Road scenes sampled from $p_{test}(x, y)$ consist of variations and combinations of structural constituents found in the training road scenes.}
  \label{fig:test_dist}
\end{figure}

The dataset for this work is artificially generated from a set of 21 images representing variations of various symmetric and asymmetric road layouts visualized in Fig.~\ref{fig:train_dist}-\ref{fig:test_dist}. The model is trained on 75 training samples, each generated from a manually drawn feasible trajectory over an image. The training dataset consists of 2 intersections, 3 straights, 1 curve, 3 T-intersections, 2 one-way intersections, 1 Y-intersection, and 1 roundabout. A training sample is shown in Fig.~\ref{fig:training_sample}.

The test dataset consists of 8 variations of the above layouts with a different number of lanes or connections. Data can be collected by pre-processing sensor data into a top-down semantic representation of the road scene context \cite{seo19}, and superimposing recorded example trajectories obtained either from ego-vehicle odometry or tracking other vehicles. Artificial data can be generated by manually drawing example trajectories on top of a semantic road scene image as done in this work. It is important to note that the model learns to output all feasible trajectories while only observing one at a time during training.

\subsection{Data augmentation}

Data augmentation is an essential component of the training process and critical for training models which are invariant to precise road scene geometries and thus able to generalize beyond the dataset. Every training sample is randomly augmented when loaded and thus the model is trained on a theoretically infinite set of unique samples. Data augmentation is performed by random rotation and applying component-wise polynomial warping \cite{glasbey98} to the road scene context and trajectory label, mapping points in the warped sample space $i'$ to the unwarped sample space $i$ according to the following nonlinear function

\begin{equation} \label{eq:poly_warping}
 a_0 (i')^2 + a_1 i' + a_2 = i
\end{equation}

with the following boundary conditions: $i' = 0 \land i = 0$, $i' =  I_{max} \land i = I_{max}$, $i' = i'_0 \land i = i_0$. The input dimension is denoted $I_{max}$. The warp is defined by setting $(i'_0, i_0)$. The coefficients in (\ref{eq:poly_warping}) are derived using the previous boundary conditions

\begin{equation} \label{eq:poly_warping_coeff}
 a_0 = \frac{1 - a_1}{I_{max}}, \: a_1 = \frac{i_0 - (i'_0)^2 / I_{max}}{i'_0 (1 - i'_0 / I_{max})}, \: a_2 = 0
\end{equation}

In this work, $(i_0, j_0)$ are set to the context mid-point, and the warping location $(i'_0, j'_0)$ sampled from a radial Gaussian distribution with a mean centered at radius $0.15 \: I_{max}$ with values above $0.3 \: I_{max}$ clipped.

\subsection{Performance metrics}
\label{sec:performance_metric}

The loss terms (\ref{eq:sla_loss}), (\ref{eq:da_loss}) are not suitable as performance evaluation metrics as their magnitudes are hyperparameter dependent, in addition to not reflecting the overal performance on the entire road scene beyond the sample's single label trajectory. Therefore, a set of separate evaluation samples containing the set of all feasible directional lanes are generated to evaluate performance. The samples can be generated either artificially or by superposing several observed trajectories in the same road scene. Each evaluation sample consists of one layer representing the soft lane affordance output $\hat{y}^{eval}_{i,j} \in \hat{Y}^{eval}$ and directional modes $\hat{\mu}^{eval}_{m,i,j}, m \in (1, \dots, M)$  for capturing the frequency-invariant directionality at each grid point.

Performance is evaluated by feeding in the evaluation sample road context tensor to the model, and comparing the output with the evaluation label. Soft lane affordance is evaluated using cross entropy. To allow comparison between different $\alpha_{SLA}$ magnitudes, the output intensity is normalized by scaling the soft lane affordance values $y_{i,j}$ so that the highest value becomes 1 and the lowest value 0.

\begin{equation} \label{eq:eval_sla}
  L^{eval}_{SLA} = \frac{1}{n} \sum_{i,j} - \hat{y}^{eval}_{i,j} \: log \: y_{i,j} - (1 - \hat{y}^{eval}_{i,j}) \: log (1 - y_{i,j} )
\end{equation}

Directional affordance is evaluated by generating a multimodal directional distribution $\hat{p}(\theta)^{eval}_{i,j}$ using $\hat{\mu}^{eval}_{m,i,j}$ similar to Sec.~\ref{sec:da_loss}, and comparing it with the model output distribution $p(\theta)_{i,j}$ using average KL divergence (\ref{eq:kl_div}) value over all directional grid points $\tilde{n}$ as the evaluation metric.

\begin{equation} \label{eq:eval_da}
 L^{eval}_{DA} = \frac{1}{\tilde{n}} \sum_{i,j} D_{KL} (\tilde{p}(\theta)^{eval}_{i,j} \| p(\theta)_{i,j})
\end{equation}

\section{RESULTS}
\label{sec:results}

\begin{figure}
  \centering
  \includegraphics[width=0.38\textwidth]{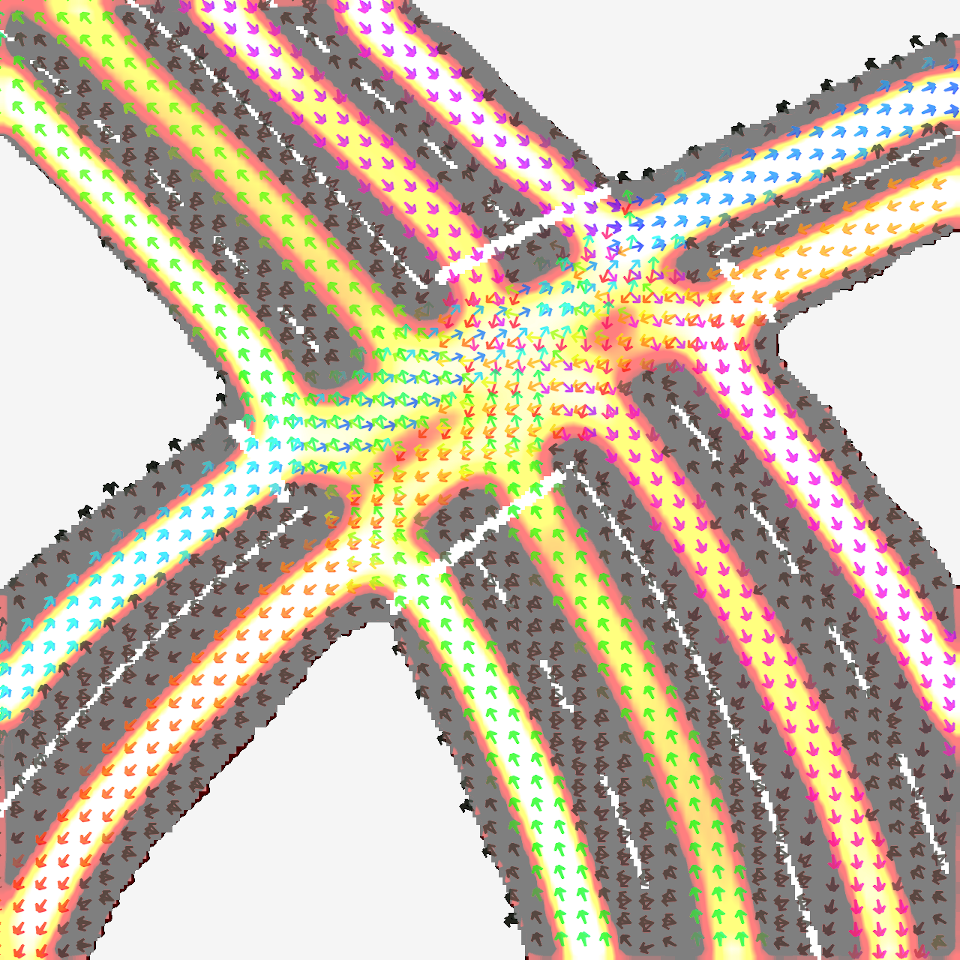}
  \caption{Visualization of model output (Exp 6) on an intersection sample generated from the training data distribution $p_{data}(x,y)$. The heat map corresponds to the predicted likelihood of a grid point belonging to a soft lane. Each arrow correspond to a directional mode, defining the predicted multimodal directionality of traffic at the point.}
  \label{fig:train_visualization}
\end{figure}

\begin{figure}
  \centering
  \includegraphics[width=0.48\textwidth]{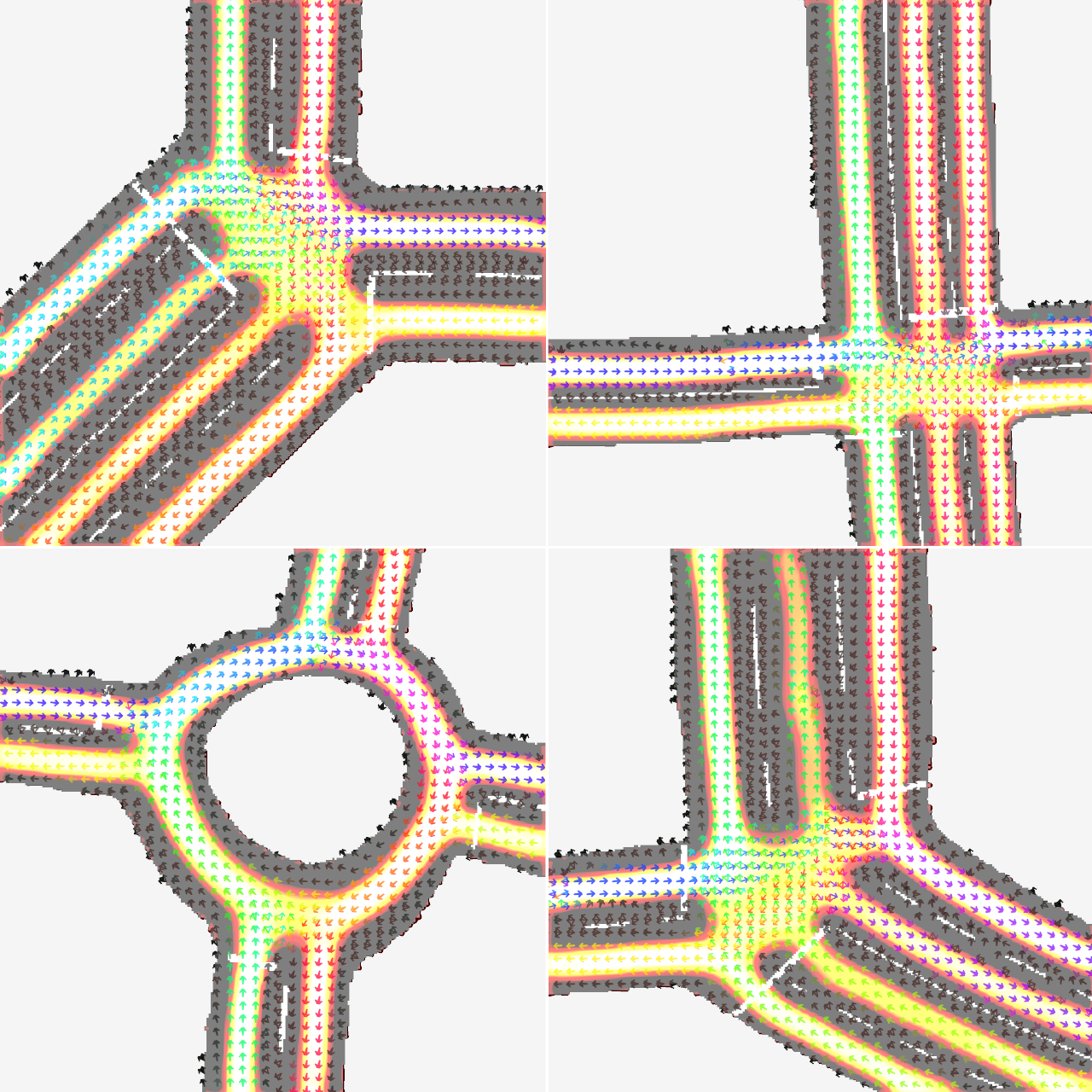}
  \caption{Examples demonstrating generalizability. The model (Exp 6) has learned a useful representation of the road, and correctly infers directional road lanes for new road scenes sampled from $p_{test}(x,y)$.}
  \label{fig:test_visualization}
\end{figure}

All experimental results are summarized in Table~\ref{tab:summary}, where $\eta$ denotes learning rate, $p_{drop}$ is the dropout probability, and $\alpha_{SLA}$ is the mask loss strength. The remaining four values show the evaluation metric on the training and testing sample distribution. A set of ten randomly generated samples for each road layout (see Fig.~\ref{fig:train_dist}-\ref{fig:test_dist}) are used for evaluation and visualizations. Fig.~\ref{fig:train_visualization} shows a visualization of an output generated from $p_{data}(x,y)$, and Fig.~\ref{fig:test_visualization} displays the correct road lanes and multimodal directionality for four road layouts the model has never encountered during training, demonstrating that the model has learned a useful representation and can generalize to new road layouts and geometries. All models were trained for 2500 epochs and reached convergence. One training run takes approximately 8 hours on an Intel i9-9900K CPU and Nvidia RTX 2080 Ti GPU.

\begin{table*}[h]
\caption{Summary of experiments and results.}
\label{tab:summary}
\begin{center}
\begin{tabular}{|c|c|c|c|c|c|c|c|c|}
\hline
Exp & $\eta$ &$p_{drop}$ & $\alpha_{SLA}$ & $L^{eval}_{SLA, train}$ & $L^{eval}_{DA, train}$ & $L^{eval}_{SLA, test}$ & $L^{eval}_{DA, test}$\\
\hline
1 & $6\mathrm{e}^{-6}$ & 0.2 & 100 & $\boldsymbol{0.259}$ & 1.294 & 0.297 & 0.338 \\
\hline
2 & $6\mathrm{e}^{-6}$ & 0.2 & 10 & 0.357 & 1.256 & 0.342 & 0.345 \\
\hline
3 & $6\mathrm{e}^{-6}$ & 0.2 & 1 & 0.412 & 1.168 & 0.362 & 0.358 \\
\hline
4 & $9\mathrm{e}^{-6}$ & 0.2 & 100 & 0.265 & 1.265 & 0.332 & 0.539 \\
\hline
5 & $3\mathrm{e}^{-6}$ & 0.2 & 100 & 0.278 & 1.237 & 0.323 & 0.529 \\
\hline
6 & $6\mathrm{e}^{-6}$ & 0.0 & 100 & 0.264 & $\boldsymbol{1.059}$ & $\boldsymbol{0.292}$ & $\boldsymbol{0.319}$ \\
\hline
7 & $6\mathrm{e}^{-6}$ & 0.4 & 100 & 0.260 & 2.626 & 0.303 & 0.909 \\
\hline
\end{tabular}
\end{center}
\end{table*}

\begin{figure}
  \centering
  \includegraphics[width=0.40\textwidth]{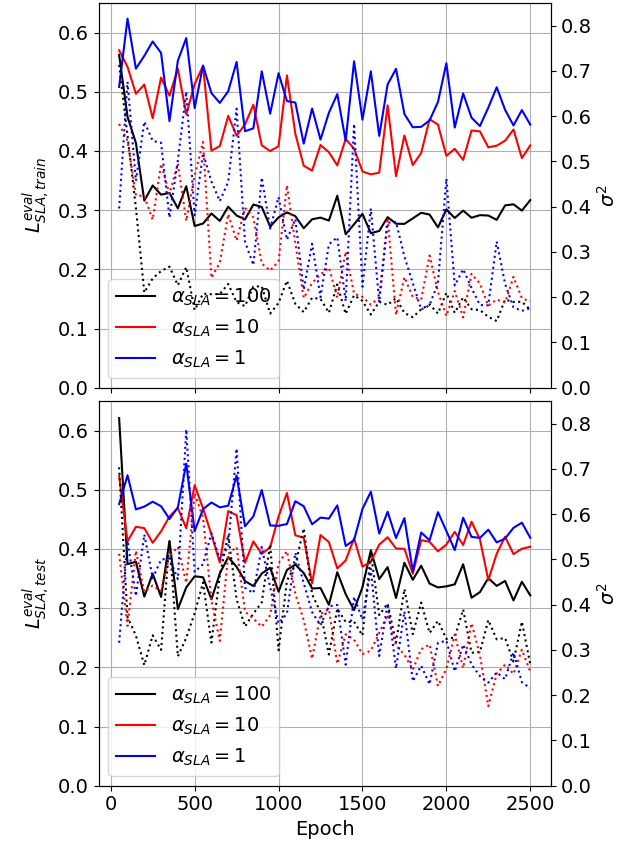}
  \caption{Learning performance on the training evaluation samples (upper) and the test evaluation samples (bottom) with varying $\alpha_{SLA}$. The solid line and dashed lines represent the mean and variance respectively.}
  \label{fig:sla_comparison}
\end{figure}

The effect of varying the soft lane affordance masking value $\alpha_{SLA}$ is seen in Fig.~\ref{fig:sla_comparison}. A larger value results in the best performance on both the train and test distribution. It is believed that the larger impact of the elements masked by the trajectory label encourage the model to learn a richer representation of the road scenes. The directional affordance varied negligibly across the experiments.

\begin{figure*}
\centering
\begin{minipage}{.48\textwidth}
  \centering
  \includegraphics[width=0.84\linewidth]{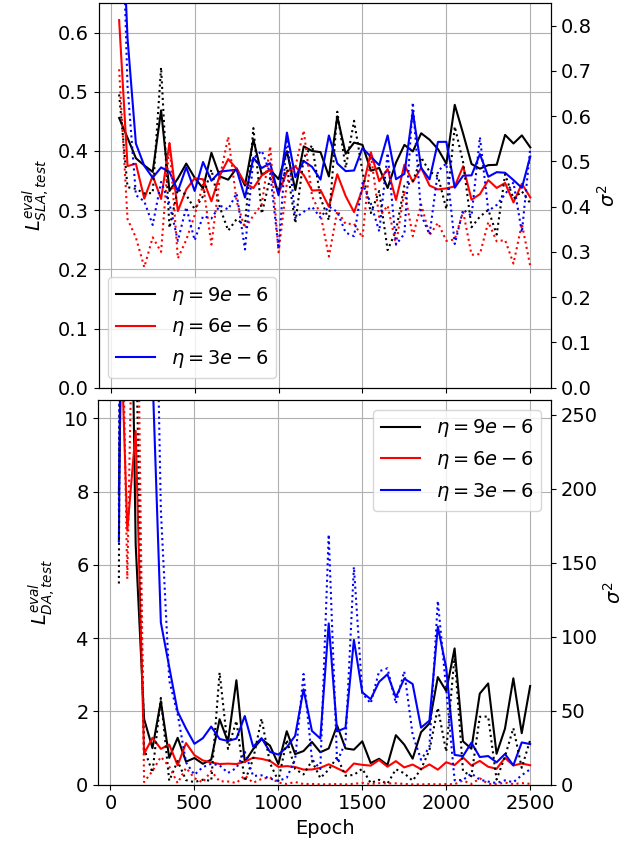} 
  \caption{Learning performance on the test evaluation samples with varying learning rates $\eta$. The upper figure shows soft lane accuracy and the lower one shows directional accuracy. The solid line and dashed lines represent the mean and variance respectively.}
  \label{fig:lr_comparison}
\end{minipage}%
\begin{minipage}{.04\textwidth} \hspace{\fill}
 \end{minipage}%
\begin{minipage}{.48\textwidth}
  \centering
  \includegraphics[width=0.84\linewidth]{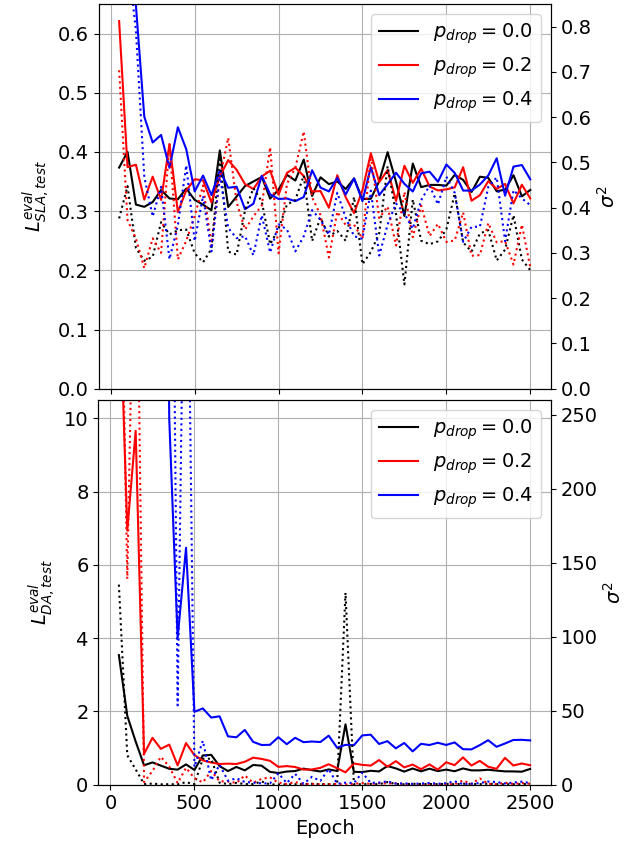}
  \caption{Learning performance on the test evaluation samples with varying dropout rates $p_{drop}$. The upper figure shows soft lane accuracy and the lower one shows directional accuracy. The solid line and dashed lines represent the mean and variance respectively.}
  \label{fig:dropout_comparison}
\end{minipage}
\end{figure*}

An initial learning rate of $\eta = 6e^{-6}$ results in the best learning performance on both the train and test distributions. In particular the directional performance on the test distribution is notably more stable. The test distribution performance is shown in Fig.~\ref{fig:lr_comparison}.

Dropout is found to decrease generalization performance as seen in Fig.~\ref{fig:dropout_comparison}. The reason is believed to be that every training example is unique and drawn from the true training distribution $p_{train}(x, y)$, meaning that the model does not overfit the training dataset and therefore dropout only hampers the learning process by introducing noise.

\begin{figure}
  \centering
  \includegraphics[width=0.48\textwidth]{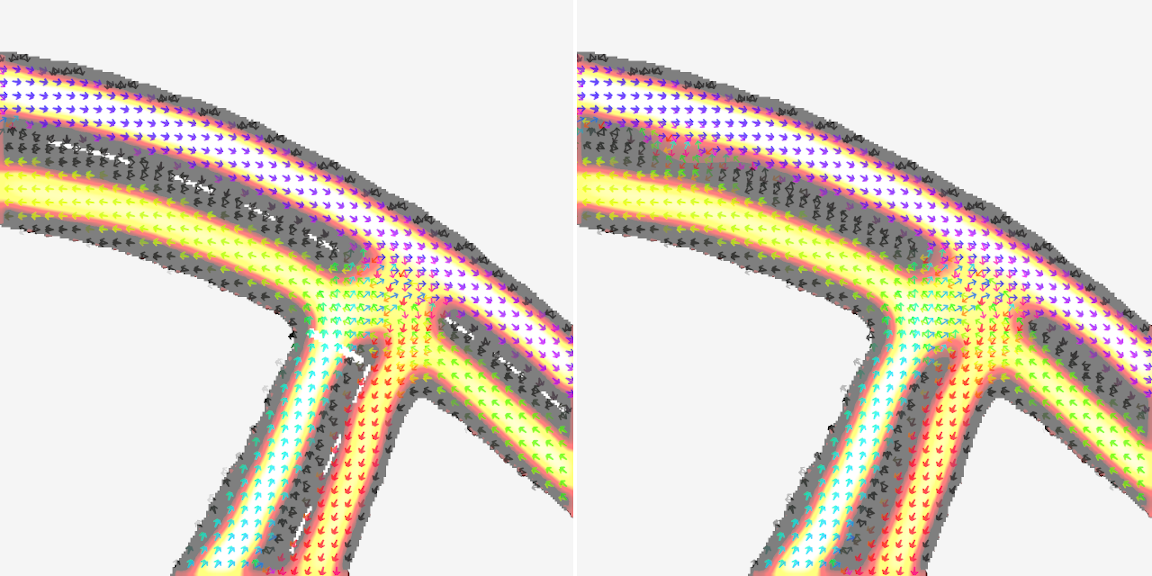}
  \caption{Visualization of feature independence. The left and right images show equivalent output for the same intersection both with and without road markings.}
  \label{fig:feature_independence}
\end{figure}

An important property of the CNN-based model is feature independence. Fig.~\ref{fig:feature_independence} illustrates how the model is able to learn a representation which is not dependent on a single set of features by being trained on samples with and without road markings, and thus is more robust than conventional semi-automatic HD map generation methods \cite{guo16} which break down for road scenes without road markings etc.

Observed failure modes include noisy directions because the model's normalized directional mean output range $\tilde{\mu}_m \in [0, 1]$ causes clipping for regions where angles shift between $0^{\circ}$ and $360^{\circ}$. This problem is believed to be resolvable by changing the directional output representation from a scalar to a biternion representation \cite{beyer15}. Another observed failure mode occurs on one-way streets, resulting in the model losing the sense of directionality far away from the intersection. The reason is believed to be lack of informative local contextual features in the vicinity and fading contextual information originating from the intersection region. Supplementing the dense model with a graphical model \cite{newell17, yang18} and/or applying iterative reasoning \cite{cao17} is believed to resolve the problem through improved global contextual reasoning.

\section{CONCLUSION}
\label{sec:conclusion}

This work presents a novel self-supervised method for learning a probabilistic network model to infer paths or road lanes with direction conditioned on contextual features of the environment. The model can be trained to output all feasible road lanes with direction learning only from examples containing only one trajectory label at a time. Data for training can be automatically collected from driving data. A randomized rotation and warping algorithm is used to augment training and test data, resulting in an online learning method where the model is directly trained and evaluated on a data distribution. Training performance is quantitatively analyzed over a range of hyperparameters and generalization ability is demonstrated quantitatively and visually by successfully inferring the correct directional road lane structure on road layouts not encountered during training. Additionally, the proposed model is shown to be robust against feature dependence, unlike other methods.

Generalization ability can be further motivated by noting that the model is shown to learn a spatial invariance to towards precise geometry of a road layout, and given that there exist a finite set of feasible road layouts \cite{fhwa17}, learning directional road lanes becomes to be tractable problem.

Future work includes improving the model's global contextual reasoning capability, as well as evaluating the model using real-world data collected from onboard sensors and/or a public dataset like Cityscapes~\cite{cordts16}. Previous similar works \cite{salzmann19, baumann18, barnes17, perezhigueras18} have demonstrated that CNN-based contextual multimodal path prediction approaches can work with noisy and partially occluded real-world input representations, and thus we expect our method will extend to these input feature types.

\end{document}